\documentclass[sigconf]{acmart}
\AtBeginDocument{%
  }

\copyrightyear{2024}
\acmYear{2024}
\setcopyright{acmlicensed}\acmConference[CIKM '24]{Proceedings of the 33rd
ACM International Conference on Information and Knowledge
Management}{October 21--25, 2024}{Boise, ID, USA}
\acmBooktitle{Proceedings of the 33rd ACM International Conference on
Information and Knowledge Management (CIKM '24), October 21--25, 2024,
Boise, ID, USA}
\acmDOI{10.1145/3627673.3679852}
\acmISBN{979-8-4007-0436-9/24/10}

\usepackage{graphicx}
\usepackage{enumitem}
\usepackage{booktabs}
\usepackage{multirow}
\usepackage{tabularx} 
\begin{document}

\title{M\textsuperscript{2}ConceptBase: A Fine-Grained Aligned Concept-Centric Multimodal Knowledge Base}

\author{Zhiwei Zha}
\orcid{0009-0002-7005-0372}
\affiliation{%
  \institution{Shanghai Key Laboratory of Data
Science, School of Computer Science, Fudan University}
  \city{Shanghai}
  \country{China}
}
\email{zwcha22@m.fudan.edu.cn}

\author{Jiaan Wang}
\affiliation{%
  \institution{School of Computer Science and
Technology, Soochow University}
  \city{Suzhou}
  \country{China}}
\email{jawang.nlp@gmail.com}

\author{Zhixu Li}
\authornotemark[1]
\affiliation{%
 \institution{Shanghai Key Laboratory of Data
Science, School of Computer Science, Fudan University}
 \city{Shanghai}
 \country{China}
}
\email{zhixuli@fudan.edu.cn}

\author{Xiangru Zhu}
\affiliation{%
 \institution{Shanghai Key Laboratory of Data
Science, School of Computer Science, Fudan University}
 \city{Shanghai}
 \country{China}
}
\email{xrzhu19@fudan.edu.cn}

\author{Wei Song}
\affiliation{%
 \institution{Research Center for Intelligent Robotics, Zhejiang Lab}
 \city{Hangzhou}
 \country{China}
}
\email{weisong@zhejianglab.com}

\author{Yanghua Xiao}
\affiliation{%
  \institution{Shanghai Key Laboratory of Data
Science, School of Computer Science, Fudan University}
  \city{Shanghai}
  \country{China}
}
\email{shawyh@fudan.edu.cn}


\begin{abstract}

Multimodal knowledge bases (MMKBs) provide cross-modal aligned knowledge crucial for multimodal tasks. However, the images in existing MMKBs are generally collected for entities in encyclopedia knowledge graphs. Therefore, detailed groundings of visual semantics with linguistic concepts are lacking, which are essential for the visual concept cognition ability of multimodal models. Addressing this gap, we introduce M\textsuperscript{2}ConceptBase, the first concept-centric MMKB. M\textsuperscript{2}ConceptBase models concepts as nodes with associated images and detailed textual descriptions. We propose a context-aware multimodal symbol grounding approach to align concept-image and concept-description pairs using context information from image-text datasets. Comprising 951K images and 152K concepts, M\textsuperscript{2}ConceptBase links each concept to an average of 6.27 images and a single description, ensuring comprehensive visual and textual semantics. Human studies confirm more than 95\% alignment accuracy, underscoring its quality. Additionally, our experiments demonstrate that M\textsuperscript{2}ConceptBase significantly enhances VQA model performance on the OK-VQA task. M\textsuperscript{2}ConceptBase also substantially improves the fine-grained concept understanding capabilities of multimodal large language models through retrieval augmentation in two concept-related tasks, highlighting its value.\footnote{The data and codes are available at  https://github.com/AwellmanZha/M2ConceptBase}

\end{abstract}




\maketitle

\begin{figure}[t]
\centerline{\includegraphics[width=0.85\columnwidth]{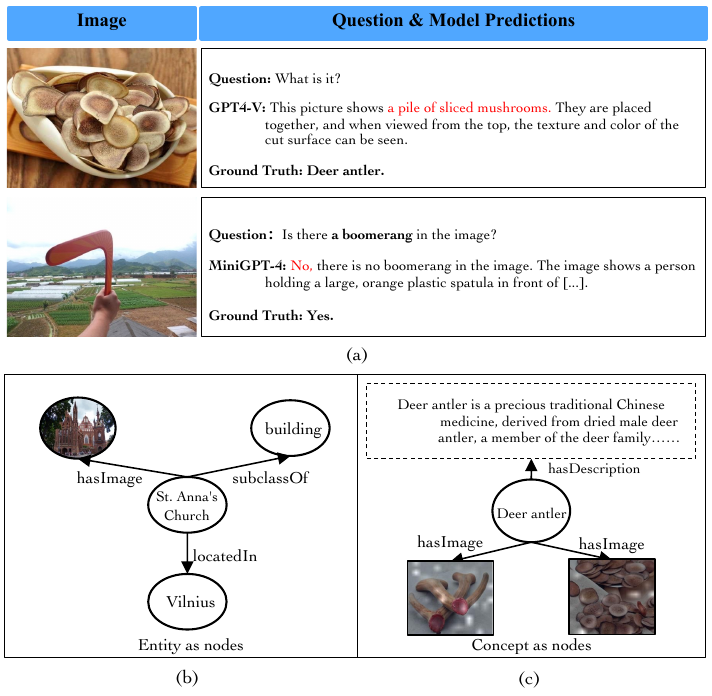}}
\caption{(a) Examples of sub-optimal fine-grained concept understanding in GPT-4V and miniGPT-4, highlighting the need for concept-centric MMKBs. (b) Entity-centric MMKBs. (c) Concept-centric MMKBs.}
\label{fig:intro}
\end{figure}

\begin{table*}[t]
\centering
\resizebox{0.95\textwidth}{!}{
\begin{tabular}{lllllll}
\toprule[1pt]
\multicolumn{1}{c}{MMKB} & \multicolumn{1}{c}{Characteristics} & \multicolumn{1}{c}{Scale (\#nodes/\#images)} & \multicolumn{1}{c}{Image Grain} &  \multicolumn{1}{c}{Data Source} \\ \midrule[1pt]
VisualSem~\cite{alberts2020visualsem}                & Entity-centric KB     & 90K(entities)/938K  &   Corsed-grained                      & Wikipedia, WordNet, ImageNet        \\

MMKG~\cite{liu2019mmkg}                     & Entity-centric KB    &     45K (entities)/37K  &     Corsed-grained              & Freebase, DBpedia, YAGO, Image Search Engine  \\
Richpedia~\cite{wang2020richpedia}                 &Entity-centric KB     & 2.8M (entities)/2.9M  &    Corsed-grained      &   Wikidata, Wikimedia, Image Search Engine            \\ 

IMGpedia~\cite{ferrada2017imgpedia}  &Entity-centric KB     & 2.6M (entities)/15M    &    Corsed-grained                     &   Wikimedia Commons, DBpedia            \\ 

ImageGraph~\cite{liu2017robust}  &Entity-centric KB     & 15K (entities)/837K  &    Corsed-grained       &  Freebase, Image Search Engine            \\ 

ImageNet~\cite{russakovsky2015imagenet}                 &  Image Dataset  & 21K (classes)/3.2M   & Corsed-grained                           & WordNet, Image Search Engine          \\

NEIL~\cite{chen2013neil}             &     Image Database   & 1152 (classes)/300K        &       Fine-grained                                         & WordNet, Image Search Engine       \\

GAIA~\cite{li2020gaia}             &   Entity-centric KB    & 457K (entities)/NA   &       Fine-grained                                         & Freebase, GeoNames, Multimeida News Websites      \\

RESIN~\cite{wen2021resin}             &   Event-centric KB    &  51K (events)/NA  &       Corsed-grained                                         & Wikidata, Multimeida News Websites      \\

VisualGenome~\cite{krishna2017visual}             &     Image Dataset   & 35 (classes)/108K  &       Corsed-grained                                         & WordNet, MS COCO, YFCC       \\ 
\midrule[1pt]
\textbf{M\textsuperscript{2}ConceptBase}                & \textbf{Concept-centeric KB}          &   \textbf{152K(concepts\&descriptions)/951K} & \textbf{Fine-grained}       & \textbf{Image-text Pairs, Encyclopedia}  \\
\bottomrule[1pt]
\end{tabular}
}
\caption{Key features of M\textsuperscript{2}ConceptBase compared to other multimodal knowledge bases.} %
\label{tab:statistics}
\end{table*}

\begin{figure*}[t]
\centerline{\includegraphics[width=1.8\columnwidth]{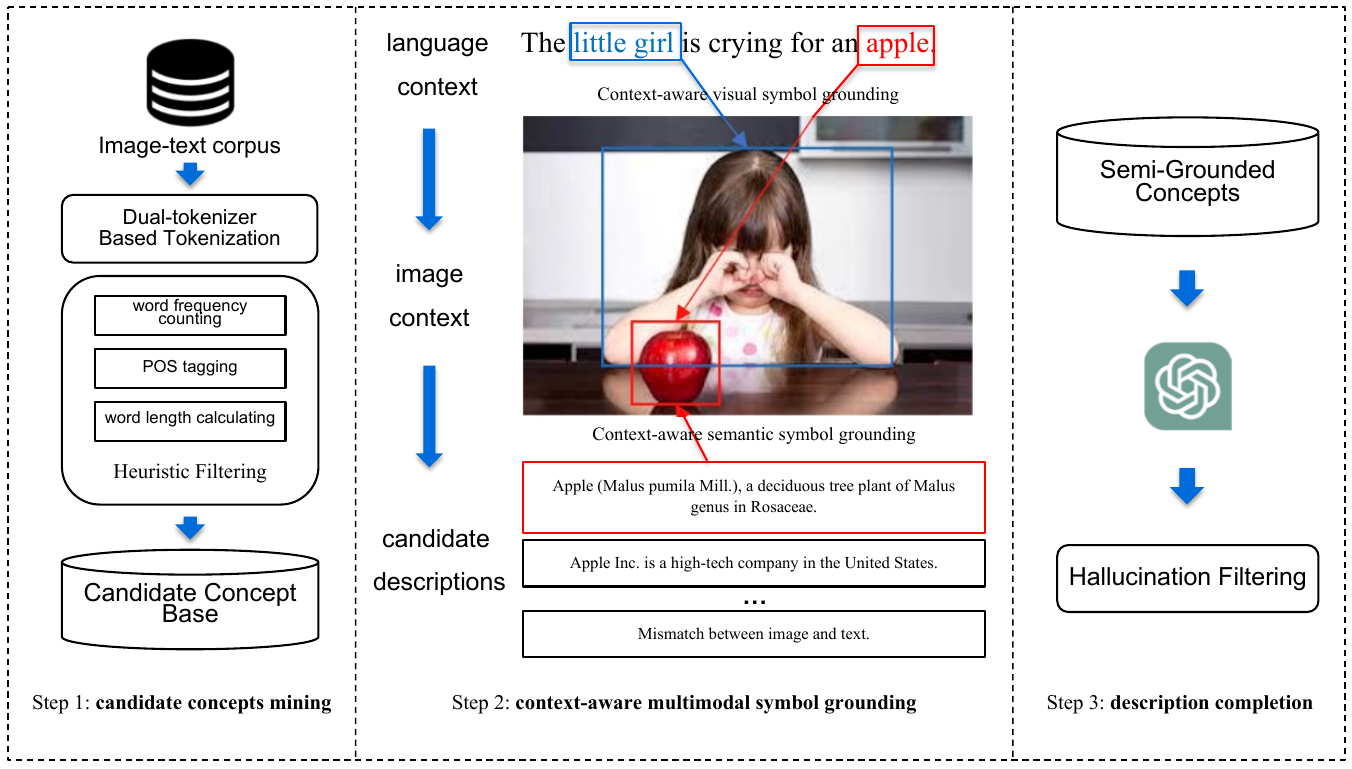}}
\caption{Our framework for large-scale concept-centric multimodal knowledge base construction. In step 1, we mine candidate concepts from large-scale image-text pairs by tokenizing their textual descriptions and filtering the tokenized results by rule-based strategies. In step 2, we ground each candidate concept with concept-relevant images and detailed concept descriptions. In step 3, we generate concept descriptions for those concepts that failed to be grounded in step 2.}
\label{fig:framework}
\end{figure*}

\section{Introduction}



Multimodal knowledge bases (MMKBs) integrate various types of information, \emph{e.g.}, textual and visual data, to achieve a comprehensive understanding of the world. This comprehensive representation enhances the cognitive abilities of artificial intelligence systems and provides valuable resources for numerous downstream tasks, such as visual recognition, knowledge-based visual question answering, reasoning, and robotics~\cite{wu2023recognizing, wang2023vqa, zhu2022multimodal, zha2022exploring, li2023kerm}. 
Furthermore, although Multimodal Large Language Models (MLLMs) have shown promising results in various multimodal tasks, they still exhibit sub-optimal performance in fine-grained concept understanding (as shown in Figure ~\ref{fig:intro}(a)) in knowledge-intensive scenarios~\cite{kandpal2023large, liu2023mmbench, xu2023lvlm}. Existing works have attempted to integrate cross-modal aligned knowledge  from MMKBs into MLLMs, which have shown promising improvements in tasks requiring retrieval or knowledge augmentation~\cite{yu2021ernie,  ye2023improving, chen2023lion, yang2024give, liu2024rar, li2024kenet, li2024generative, qiu2024snapntell}. 

However, existing MMKBs~\cite{alberts2020visualsem, liu2019mmkg, wang2020richpedia, ferrada2017imgpedia, liu2017robust} generally focus on entities (named entity-centric MMKBs), leaving a gap in the exploration of concept-centric ones. Figure~\ref{fig:intro}(a) shows examples where two advanced MLLMs(\emph{i.e.} GPT-4V~\cite{OpenAI2023GPT4TR} and miniGPT-4~\cite{zhu2023minigpt}) failed in fine-grained concept understanding, highlighting the importance of cross-modal concept-aligned knowledge. Figure~\ref{fig:intro}(b) and Figure~\ref{fig:intro}(c) illustrate the differences between entity-centric and concept-centric MMKBs. Concept-centric MMKBs emphasize visual concepts, which are more informative and crucial for the scene understanding capabilities of MLLMs.
%
%
We find that existing MMKBs either rely on knowledge graphs whose nodes typically represent entities, or a knowledge extractor with predefined categories (c.f. Table~\ref{tab:statistics}).
As a result, the knowledge graph sources and limited categories restrict the acquisition of conceptual knowledge.
%

In this work, we propose to construct the first Concept-Centric Multimodal Knowledge Base with a more comprehensive coverage of visual concepts. 
However, this effort faces significant challenges in both \textbf{gathering a wide range of visualizable concepts~\cite{jiang2022visualizable} (C1)} and encountering \textbf{scarcity of directly associated images (C2)}. Previous methods of constructing entity-centric MMKBs are not suitable for building concept-centric MMKBs. Specifically, there are two paradigms for constructing traditional MMKBs:
(1) ``From Text to Image'' methods~\cite{ferrada2017imgpedia, liu2017robust, wang2020richpedia, alberts2020visualsem, zhang2023aspectmmkg} collect images for textual knowledge bases. They search for candidate images based on textual entities or concepts and then apply a filtering mechanism to obtain matched images.
(2) ``From Image to Text'' methods~\cite{chen2013neil, li2020gaia, wen2021resin, krishna2017visual} extract visual knowledge from images. They utilize automated techniques such as object detection, image tagging, and visual relationship extraction to extract entity and concept information. Alternatively, manual efforts may be employed to annotate the knowledge bases.
However, neither of these methods is suitable for our task. On one hand, ``From Text to Image'' methods require predefined concepts, and the concepts in our existing textual knowledge bases may not always be of high quality. Many of them are non-visualizable~\cite{jiang2022visualizable} and have limited relevance in multimodal contexts. On the other hand, ``From Image to Text'' methods are constrained by the given image collection and predefined categories of detected concepts, thereby limiting the coverage of concepts.

Therefore, we introduce a novel framework for concept-centric MMKB construction that naturally overcomes challenges by leveraging the characteristics of the image-text corpus: \textbf{C1}) \emph{Utilizing word frequency analysis in textual descriptions, we can extract common visual concepts, ensuring broad concept coverage.} \textbf{C2}) \emph{Additionally, image-text pairs provide abundant cross-modal alignments at the sentence level, where visual concepts in images often align semantically with relevant keywords in the paired text, enabling effective mining of semantic alignments at the concept level.} Specifically, our framework encompasses three steps. We first mine candidate concepts by tokenizing the textual descriptions in a large amount of existing image-text pairs.
Then, we perform a context-aware multimodal symbol grounding method to align each candidate concept with concept-aware images and a detailed concept description from image-text pairs via visual symbol grounding and semantic symbol grounding, respectively.
Finally, we leverage a cutting-edge LLM (~\emph{i.e.}, GPT-3.5-Turbo) to generate concept descriptions for the concepts that failed to be fully grounded (to the detailed descriptions). 

After that, we construct M\textsuperscript{2}ConceptBase, the first concept-centric MMKB, which provides rich alignments between concepts and images/descriptions. Additionally, we achieve fine-grained alignments between concepts and the corresponding salient image regions using attention maps. With a total of 951,089 images and 151,776 concepts, each associated with an average of 6.27 images, M\textsuperscript{2}ConceptBase serves as a valuable resource of aligned cross-modal knowledge. 
%
Experimental results demonstrate the practical significance of M\textsuperscript{2}ConceptBase in two key tasks: (1) In the OK-VQA task, M\textsuperscript{2}ConceptBase significantly enhances VQA model performance, improving visual comprehension and cognitive abilities. (2) In retrieval augmented generation tasks (zero-shot visual concept recognition and visual concept knowledge description generation), M\textsuperscript{2}ConceptBase substantially enhances the fine-grained concept understanding capabilities of multimodal large language models, highlighting its significant value and usefulness.

Our contributions can be summarized as follows:
\begin{itemize}[leftmargin=*,topsep=0pt]
\setlength{\itemsep}{0pt}
\setlength{\parsep}{0pt}
\setlength{\parskip}{0pt}
\item We construct the first concept-centric MMKB, M\textsuperscript{2}ConceptBase, containing 152K concepts along with their descriptions and 951K images.

\item We introduce a novel framework for constructing concept-centric MMKBs, which is both efficient and scalable. Specifically, our context-aware multimodal symbol grounding method achieves high accuracy in aligning concepts with images and descriptions within image-text contexts.

\item Experimental results show that M\textsuperscript{2}ConceptBase significantly improves the performance of VQA models and enhances the fine-grained concept understanding of MLLMs with retrieval augmentation, underscoring the critical value of M\textsuperscript{2}ConceptBase.
\end{itemize}

\section{M\textsuperscript{2}ConceptBase}

To construct M\textsuperscript{2}ConceptBase, we start with large-scale image-text pairs that are handy to obtain in existing multimodal pre-training data.
%
%
We devise a three-step construction framework, depicted in Figure~\ref{fig:framework}.
We first mine candidate concepts from the textual descriptions of large-scale image-text pairs (\S~\ref{subsec:3.2}). Then, we propose a context-aware multimodal symbol grounding algorithm to collect the fine-grained alignment between concepts and images, and that between images and detailed concept descriptions (\S~\ref{subsec:3.3}). Lastly, we use GPT-3.5-Turbo to supplement detailed concept descriptions for the concepts that failed to be aligned in our multimodal symbol grounding algorithm (\S~\ref{subsec:3.4}).



\subsection{Problem Formulation}

Let $\mathcal{P} = \{\mathcal{I}, \mathcal{T}\}$ be a multimodal image-text corpus, where $\mathcal{I} = \{i_1, i_2, ..., i_m\}$ represents the set of $m$ images and $\mathcal{T} = \{t_1, t_2, ..., t_m\}$ represents the set of $m$ textual descriptions. Besides, let $\mathcal{E} = \{e_1, e_2, \\..., e_n\}$ be the set of $n$ encyclopedia descriptions. Our goal is to construct a concept-centric MMKB $\mathcal{B}$ from $\mathcal{P}$ and $\mathcal{E}$. Specifically, a concept-centric MMKB $\mathcal{B} = \{\mathcal{C}, \mathcal{I_{B}}, \mathcal{E_{B}}\}$ consists of nodes representing concepts. Each concept $c \in \mathcal{C}$ is associated with corresponding visual images $\{i_{c_1}, i_{c_2}, ..., i_{c_n} | i_{c_j} \in \mathcal{I_{B}}\}$ and textual descriptions $e \in \mathcal{E_{B}}$.

\subsection{Candidate Concept Mining}
\label{subsec:3.2}

As our analyses above,  the textual descriptions $\mathcal{T}$ in the image-text corpus $\mathcal{P}$ often contain potential visual concepts (typically manifest as nouns) that correspond to the objects in the images.
The goal of candidate concept mining is to obtain general concepts $\mathcal{C}$ from the textual descriptions $\mathcal{T}$:

\begin{equation}
\mathcal{C} \gets \operatorname{ConceptMining}(\mathcal{T})
\end{equation}

To obtain candidate concepts with high recall, we retain as many candidate concepts as possible in this step and use four filtering strategies to remove irrelevant words/phrases.
Specifically, for obtaining candidate concepts, we first tokenize the textual descriptions from the large-scale corpus $\mathcal{P}$ to obtain a vast collection of words/phrases and then perform word frequency statistics and part-of-speech analysis on these words/phrases to obtain candidate concepts.

\vspace{0.5ex}
\noindent \textbf{Dual-tokenizer Based Tokenization.}
To enhance the recall rate of candidate concepts, we use a dual-tokenizer based tokenization method. In detail, we use both Jieba\footnote{\url{https://github.com/fxsjy/jieba}} and LAC~\cite{jiao2018LAC} tokenizers\footnote{Jieba tokenizer tends to produce finer and shorter phrases, while LAC tokenizer is more likely to produce semantically meaningful compound phrases.} to tokenize each textual description $t_i$ in Wukong corpus~\cite{gu2022wukong} (a Chinese image-text corpus):
\begin{equation}
\small
W_{J,t_i} = \{ (w_{j_1}, p_{j_1}), ..., (w_{j_m}, p_{j_m}) \} \gets \text{Jieba}(t_i),
\end{equation}
\begin{equation}
\small
W_{L,t_i} = \{ (w_{l_1}, p_{l_1}), ..., (w_{j_n}, p_{j_n}) \} \gets \text{LAC}(t_i),
\end{equation}
where $W_{J,t_i}$ and $W_{L,t_i}$ denote the tokenized results of $t_i$ via Jieba and LAC tokenizers, respectively. $w_{j_k}$ and $w_{l_k}$ indicate the $k$-th word in $W_{J,t_i}$ and $W_{L,t_i}$, respectively, and $p_{j_k}$ and $p_{l_k}$ are their corresponding part-of-speech (POS) tags.

Then, we integrate the results as the preliminary candidate concepts to ensure robustness when faced with complex concept relationships described in the input sentences:
\begin{equation}
\mathcal{C}_\text{pc} = \bigcup_{t \in \mathcal{T}, \rho \in \{J,L\}} W_{\rho,t}
\end{equation}

As a result, we obtain about 1.18M preliminary tokenized Chinese words/phrases, denoted as $\mathcal{C}_\text{pc}$.

\vspace{0.5ex}
\noindent \textbf{Heuristic Filtering.}
To obtain candidate concepts from the preliminary tokenized results ($\mathcal{C}_\text{pc} \rightarrow \mathcal{C}$), we make use of four rule-based filtering strategies, including POS filtering, word frequency filtering, word length filtering, and supplementary compound filtering. Since a potential candidate concept must be a noun, we utilize an off-the-shelf toolkit (\emph{i.e.}, Jieba) to calculate the POS tags for each tokenized result and filter out all non-noun words.
Further, we retain phrases with a frequency greater than or equal to fifteen as candidate concepts.
Besides, we filter out phrases with a character-level length longer than five.
Since the Chinese words/phrases might involve English abbreviation, we retain (a) all English words with the POS tag ``n''; (b) all Chinese words with the POS tag ``nz''; (c) the top-50 English words with ``nz'' POS tag; (d) high-frequency Chinese words with ``ns'', ``nt'', and ``nw'' POS tags with frequency thresholds of 3000, 400, and 300, respectively.

Through the above mining process, we ultimately obtain 573,031 concepts (denoted as $\mathcal{C}=\{c_1, c_2, ..., c_{|\mathcal{C}|}\}$) in total.


\subsection{Context-aware Symbol Grounding}
\label{subsec:3.3}
Symbol grounding refers to the process of semantically linking an abstract linguistic symbol with corresponding information from other modalities. In our scene, the multimodal symbol grounding collects the alignments between each candidate concept $c \in \mathcal{C}$ and the concept-relevant images $\{i_{c_1}, i_{c_2}, ..., i_{c_n} | i_{c_j} \in \mathcal{I}\}$, and the alignments between the concept-relevant images and the detail concept descriptions $\{t_{c_1}, t_{c_2}, ..., t_{c_n} | t_{c_j} \in \mathcal{T}\}$. In this manner, we can create images and detailed descriptions associated with the concepts.

To take the potential concept ambiguity into account, we propose a context-aware multimodal symbol grounding approach, which consists of two stages to achieve cross-modal symbol grounding of concepts. Our key insight is that concepts acquire precise meanings when placed in context. For example, given the concept ``apple'', it could refer to either the Apple company or the fruit. As shown in Figure~\ref{fig:framework} (step 2), when the concept ``apple'' appears in the context ``The little girl is crying for an apple'', along with the corresponding image, we can determine that ``apple'' refers to a fruit rather than a company. Thus, we decide to take the context information into account when performing the multimodal symbol grounding algorithm.

Specifically, we first perform \emph{visual symbol grounding} to align each candidate concept with the concept-relevant images (with attention weights based on the concept).
%
Then, we perform \emph{semantic symbol grounding} to match the weighted images with concept descriptions crawled from the encyclopedia website.



\vspace{0.5ex}
\noindent \textbf{Visual Symbol Grounding.}
Visual symbol grounding contains two sequential subprocesses: concept-activated attention-weighted image acquisition and cross-modal concept matching. 
(a) The goal of \emph{concept-activated attention-weighted image acquisition} is to obtain fine-grained attention-weighted image regions $\{\hat{i}_{c_1}, \hat{i}_{c_2}, ..., \hat{i}_{c_n} | \hat{i}_{c_j} \in \mathcal{I}\}$ activated by each concept $c$, where $\hat{i}_{c_j}$ denotes the weighted image $i_{c_j}$.
Inspired by~\citet{chefer2021generic}, we use attention mechanisms to emphasize the regions in the image that correspond to the activated concept $c$, resulting in a weighted image.
Formally, given an image-text pair $\langle i, t \rangle \in \mathcal{P}$, we tokenize the textual description $t$ and retain the concepts that appear in the candidate concepts $\mathcal{C}$, obtaining pairs of $\langle$image, concept set$\rangle$ as $\langle i, C_i = \{c_1, c_2, ..., c_k, ...\} \rangle$.
%

For each concept $c \in C_i$, we input the prompt ``\emph{an image of [concept]}'' into the text encoder of the CLIP model~\cite{radford2021learning} and the corresponding image $i$ into the vision encoder of the CLIP to obtain the output, which
denoted as $y_c$:
\begin{equation}
y_c \gets \text{CLIP}(i, \text{``an~image~of~\{\emph{concept}\}.''}) 
\end{equation}
Then, the image $i$ will be reshaped into $m\times n$ image patches.
Further, we calculate the relevance score matrix $R_i \in \mathbb{R}^{m\times n}$ by the self-attention matrix in each visual encoder layer.

Specifically, with the contextualization of tokens through attention layers, we get the relevance score matrix $R_i$:
\begin{equation}
R^{l}_i \gets R^{l-1}_i + \bar{A}_{l} \odot R^{l-1}_i, l \in \{1,2,...,L\}
\end{equation}
\begin{equation}
\bar{A}_{l} = E_h((\nabla A_{l} \odot A_{l})^+),  l \in \{1,2,...,L\}
\end{equation}
where $R^{0}_i$ is initialized with the identity matrix $I$, $R^{L}_i$ means $R_i$, which iteratively aggregating each layer’s attention weights. $L$ indicates the total number of visual layers. $A_{l}$ indicates the $i$-th layer's attention weights. $\nabla A_{l} = \frac{\partial y_c}{\partial A_{l}} $ is the concept activation gradient, $\odot$ represents the Hadamard product, $^+$ means clampping to zero to remove the negative contributions and $E_h$ means an average across self-attention heads.
%
Each element in $R_i$ denotes the relevance between each image patch of $i$ and the concept $c$. Afterward, the bilinear interpolation algorithm is applied to calculate an image weight denoted as $w_{i,c}$:
\begin{equation}
w_{i,c} \gets \text{bilinear\_interpolation}(R_i)
\end{equation}
which highlights the most relevant region of image $i$ corresponding to the target concept $c$. The weight $w_{i,c}$ is normalized (represented as $\tilde{w}_{i,c}$) and integrated back into the image pixels to emphasize important regions in the original image. This is represented as:
\begin{equation}
\hat{i}_c = \tilde{w}_{i,c} \oplus i
\end{equation}
where $\oplus$ is a pixel-wise addition operation.

%



    


(b) After obtaining the concept-activated attention-weighted images $\hat{i}$, we perform \emph{cross-modal concept matching} to only retain high-quality pairs of weighted images and concepts. Specifically, given a $c \in C_i$ and a weighted image $\hat{i}_c$ (aligned by the concept-activate attention-weighted image acquisition).
For each concept $c' \in C_i$, we calculate the matching score between $c'$ and the weighted image $\hat{i}_c$ via CLIP:
\begin{equation}\label{eq:10}
\small
   score = \text{CLIP}(\hat{i}_c, \text{``an~image~of~\{\emph{concept}\}.''}) 
\end{equation}
Only if the concept $c$ achieves the highest matching score with $\hat{i}_c$ among all concepts in $C_i$, we retain the paired $\langle$concept $c$, weighted image $\hat{i}_c\rangle$ in a semi-grounded concept base, along with the corresponding matching score. This allows us to perform sorting and obtain higher-quality paired images.

\vspace{0.5ex}
\noindent \textbf{Semantic Symbol Grounding.}
After collecting concept-image pairs, we perform semantic symbol grounding to create the alignments between the (weighted) images and the detailed concept descriptions.
To create a large-scale collection of concept descriptions, we use the candidate concepts as search terms to query Baidu Baike\footnote{\url{https://baike.baidu.com/}}, a Chinese encyclopedia. By analyzing the returned entry pages, we extract the first paragraph from the summary field as the concept description. As a concept might have multiple descriptions, we collect up to the top-3 descriptions as candidate descriptions for the concept.
After that, we successfully obtain encyclopedia descriptions for 325,925 candidate concepts.
The descriptions of concept $c$ is denoted as $t_c = \{t^1_c, t^2_c, t^3_c\}$ ($t^2_c$ and $t^3_c$ might be empty text).
To ensure the descriptions are relevant to concepts, we apply heuristic rules based on regular expressions\footnote{Regular expressions that match left and right book title marks.} to filter out non-concept descriptions, which are typically descriptions of entities.

Next, for each concept in the pairs of $\langle$weighted image $\hat{i}_c$, concept $c\rangle$, we use CLIP to match each weighted image $\hat{i}_c$ with the candidate descriptions of the concept $t_c$.
Specifically, given a weighted image $\hat{i}_c$ and the set of candidate concept descriptions $t_c = \{t^1_c, t^2_c, t^3_c\}$, CLIP produces the highest-scoring discrimination result as the final grounded concept description. The CLIP model can help us to select the most semantically fitting concept description for different images based on their visual context, \emph{i.e.}, the weighted images.
To handle the cases where no matched concept description is found among the candidate descriptions, we add an ``\emph{[unmatched]}'' tag to indicate a failure in the concept grounding process. For those candidate concept descriptions that are not grounded with the weighted images, we regard them as non-concept descriptions and discard them.


\begin{figure}[t]
\centerline{\includegraphics[width=0.85\columnwidth]{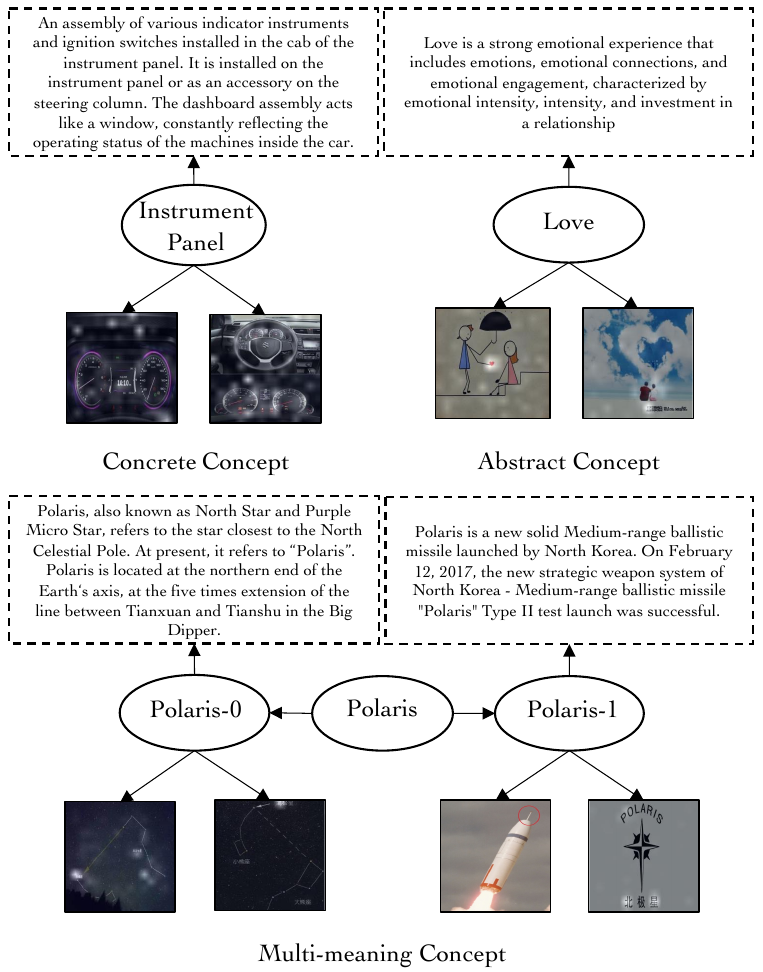}}
\caption{Example concept nodes in M\textsuperscript{2}ConceptBase.}
\label{fig:samples}
\end{figure}

\subsection{Description Completion}
\label{subsec:3.4}

After obtaining concept descriptions, there are about 233K concepts that have relevant concept descriptions, significantly less than the number of candidate concepts for pairing (about 573K, \emph{i.e.}, $|\mathcal{C}|$). As common concept descriptions are often general and abstract, which play to the strength of generative Large Language Models (LLMs), we decide to leverage GPT-3.5-Turbo~\cite{ChatGPT} to generate concept descriptions for the remaining concepts.\footnote{The used prompt is ``Please generate a basic concept description for concept \{\emph{concept}\}, scientifically and rigorously explain the basic meaning of this concept.''} 

However, we find that the preliminarily generated results of LLMs might involve a substantial amount of hallucinated content~\cite{Ji2022SurveyOH}. To address the hallucination issue, we use a simple yet effective multimodal context-based hallucination elimination mechanism, intelligently utilizing the contextual information containing concepts in the image-text pairs to eliminate hallucinated content.
%
Specifically, similar to the visual symbol grounding (\emph{i.e.}, Eq~\ref{eq:10}), the weighted images are matched with the generated concept descriptions via CLIP. In this manner, we can filter out hallucinated descriptions without any matched images. As a result, there are 87K generated descriptions, alleviating the hallucinated issue.

In addition, since the meaning of concepts is established in the textual context, we further leverage a powerful discriminative ability of LLMs to let them judge whether the concept descriptions semantically align with the concepts in the given context.
%
%
Considering the high costs of calling GPT-3.5-Turbo APIs, we decide to use another open-source powerful LLM, \emph{i.e.}, ChatGLM2-6B\footnote{\url{https://github.com/THUDM/ChatGLM2-6B}}, to make the judgment.\footnote{The used prompt is ``Context: \{\emph{text}\}; Concept: \{\emph{concept}\}; Concept description: \{\emph{description}\}; Your task is to determine whether the meaning of a concept in the Context conflicts with its description. If there is a conflict, output 0. If there is no conflict, output 1.''}
%
%
Ultimately, close to 54K out of the initial 87K concepts survive this step, successfully overcoming the issue of hallucinations in the concept description generation process by the LLMs.

\section{Data Statistics and Analyses}

In this section, we do extensive statistics and analyses to reveal the detailed features of M\textsuperscript{2}ConceptBase, including data example, detail statistics, topic coverage and data quality.

\vspace{0.5ex}
\noindent \textbf{Data Example.}
As shown in Figure~\ref{fig:samples}, the nodes in M\textsuperscript{2}ConceptBase include both concrete and abstract concepts (categorized using GPT-3.5-Turbo), and might contain multiple meanings. Each concept is accompanied by a comprehensive description and several concept-activated attention-weighted images. These attention-weighted images highlight the regions in the images that are relevant to the corresponding concepts, indicating the fine-grained alignment information provided by M\textsuperscript{2}ConceptBase.

\begin{figure}[t]
\centerline{\includegraphics[width=0.85\columnwidth]{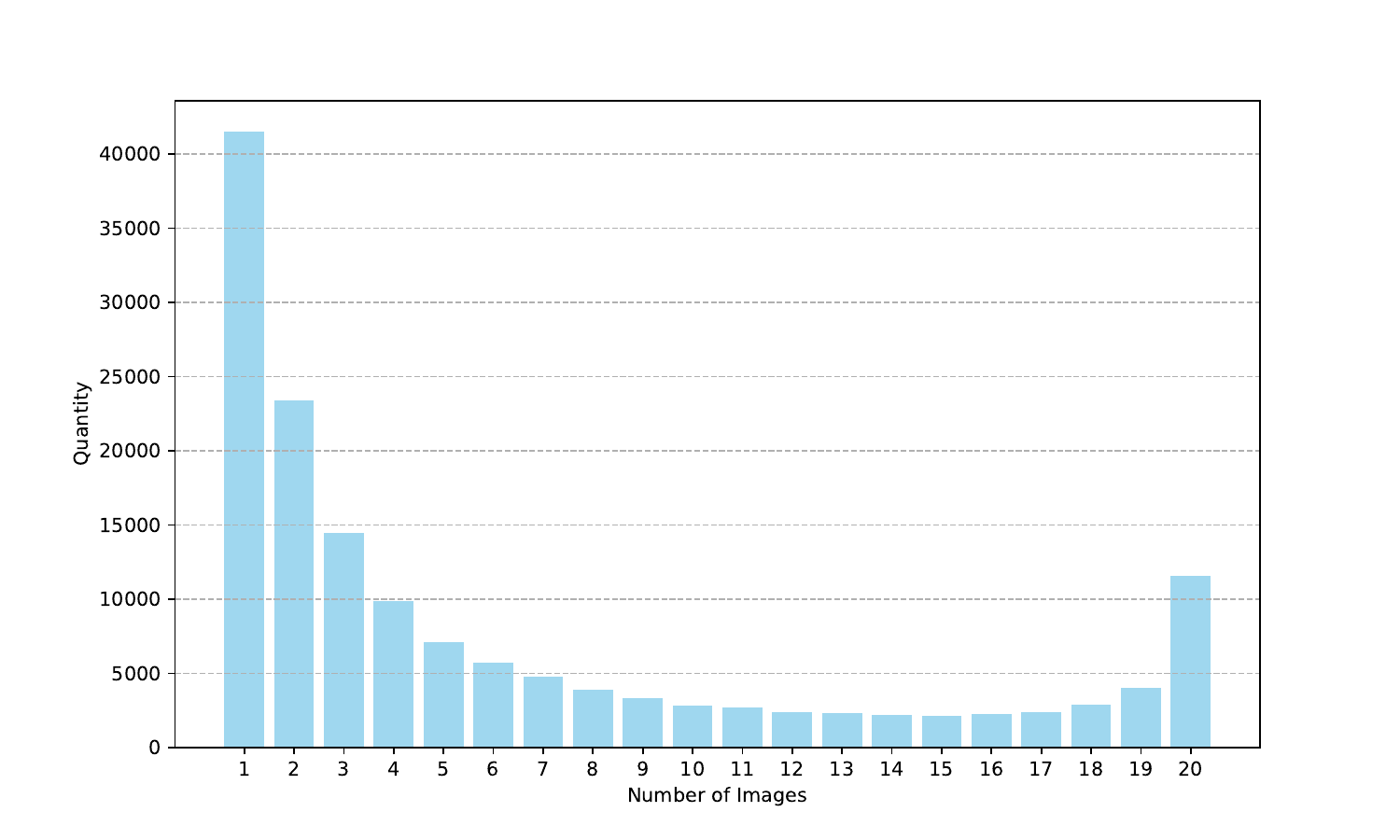}}
\caption{Distribution of the number of concepts associated with different numbers (1$\backsim$20) of images in our M\textsuperscript{2}ConceptBase.}
\label{fig:distribution}
\end{figure}

\vspace{0.5ex}
\noindent \textbf{Detail Statistics.} We introduce the detailed statistics of M\textsuperscript{2}Concept-Base as follows: M\textsuperscript{2}ConceptBase consists of 151,776 multimodal grounded concepts, each associated with multiple fine-grained weighted images activated by concepts, as well as concept description information crawled from encyclopedic knowledge sources. Each concept in M\textsuperscript{2}ConceptBase is associated with 6.27 images on average, totaling 951,089 images. M\textsuperscript{2}ConceptBase includes polysemous concepts, with 21,345 concepts containing more than one meaning, and each meaning is accompanied by a high-quality concept description text crawled from encyclopedic sources, with an average length of 105 words, containing rich concept-related knowledge.
Figure~\ref{fig:distribution} further shows a detailed distribution of the number of concepts associated with different numbers (\emph{i.e.}, 1$\backsim$20) of images. We can observe that at least 15K concepts have more than 15 images, and around 20K concepts have more than 10 images, indicating the rich fine-grained alignments provided by M\textsuperscript{2}ConceptBase.
%


\begin{table}
\centering
\small 
\resizebox{0.85\columnwidth}{!}
{
\begin{tabular}[width=0.85\columnwidth]{l|l}
\toprule[1pt]
\multicolumn{1}{c}{Topic} & \multicolumn{1}{c}{Example Concepts} \\
\midrule[1pt]
Food & Luosifen, Barbecue, Seafood, Hotpot \\
Art & Painting, Sculpture, Music, Dance \\
Health & Fitness, Nutrition, Medical, Health Preservation \\
Entertainment & Movies, Music, Games, Varieties \\
Travel & Scenery, Tourism, Scenic Spots, Natural Scenery \\
Education & Learning, Knowledge, Training, Curriculum \\
Transportation & Car, Train, Aircraft, Ship \\
Technology & Electronics, Computers, Networks, Intelligence \\
Sports & Football, Basketball, Tennis, Sporting Equipment \\
\bottomrule[1pt]
\end{tabular}
}
\caption{Example concepts w.r.t different topics.}
\label{table:example_concepts}
\end{table}


\vspace{0.5ex}
\noindent \textbf{Topic Coverage.} 
We leverage the concept description from M\textsuperscript{2}Con-ceptBase to train a topic classification model (\emph{i.e.}, Latent Dirichlet Allocation).
As shown in Table~\ref{table:example_concepts}, the results demonstrate a broad spectrum of topics covered in our M\textsuperscript{2}ConceptBase, including food, art, health, entertainment, travel, education, transportation, technology, sports, and many others.
%
%

\begin{table}[t]
\centering
\resizebox{0.85\columnwidth}{!}
{
\begin{tabular}{l|c|c}
\toprule[1pt]
\textbf{Data Quality Dimension} & \textbf{Subset} &\textbf{Accuracy(\%)} \\
\midrule[1pt]
     & concrete & \textbf{95.6} \\
Concept Confidence & abstract &\textbf{95.4} \\
     & overall & \textbf{95.5} \\
\hline
     & concrete &\textbf{96.2}\\ 
Concept-Description Alignment & abstract & \textbf{95.2}\\ 
     & overall & 
\textbf{95.9}\\ 
\hline
     & concrete &\textbf{97.7}\\ 
Concept-Image Alignment & abstract & \textbf{97.2}\\ 
     & overall & \textbf{97.5}\\  
\bottomrule[1pt]
\end{tabular}
}
\caption{Data Quality}
\label{table:dataquality}
\end{table}

\begin{table}[t]
\centering
\resizebox{0.85\columnwidth}{!}{
\begin{tabular}{l|c|c|c|c}
\toprule[1pt]
\textbf{Stage} & \textbf{Subset} & \textbf{Num. of Samples} & \textbf{Num. of Errors} &\textbf{Error Rate(\%)} \\
\midrule[1pt]
     & concrete & 100 & 3 & 3.0\\
First-check & abstract & 51 & 8 & 15.7\\ 
     & overall & 488 & 17 & 3.5\\ 
\hline
     & concrete & 100 & 1 & 1.0\\
Second-check & abstract & 50 & 4 & 8.0\\ 
     & overall & 407 & 11 & 2.7\\ 
\bottomrule[1pt]
\end{tabular}
}
\caption{The grounding accuracy of double-check mechanism.}
\label{table:grounding}
\end{table}

\vspace{0.5ex}
\noindent \textbf{Data Quality.} To assess the quality of M\textsuperscript{2}ConceptBase, we employed a crowd-sourcing strategy to determine the confidence level of grounded concepts and the accuracy of concept-description and concept-image alignments. As illustrated in Table~\ref{table:dataquality}, the results reveal an overall concept confidence of 95.5\%, indicating the percentage of concepts classified as concepts by the evaluators. Moreover, we achieved a concept-description alignment accuracy of 95.9\% and a concept-image alignment accuracy of 97.5\%, underscoring the high quality of M\textsuperscript{2}ConceptBase. 

Additionally, as depicted in Table~\ref{table:grounding}, we conducted experiments to validate the effectiveness of our double-check mechanism, which resulted in a noticeable decrease in error rate during the process. 
%
For concept confidence, we randomly sample 0.5\% of the total number of concrete and abstract concepts (i.e., 527 and 208), and engage ten volunteers to evaluate the reliability of each concept's classification as a concept.
During the assessment, we allow volunteers to use search engines. As depicted in Table~\ref{table:dataquality}, we compute the average results from ten volunteers, and obtain concrete, abstract, and overall accuracies of 95.6\%, 95.4\%, and 95.5\%, respectively.
For the cross-modal alignment accuracy, we randomly sample 0.25\% of the total number of concrete and abstract concepts (\emph{i.e.}, 263 and 104), each paired with randomly sampled (at most) 5 grounded images and the corresponding descriptions.
We invite ten volunteers to assess the accuracy of concept-image pairing by determining the number of correctly matched images in the sampled set.
Additionally, volunteers evaluate concept-description pairing based on the instruction ``\emph{Does the text correctly describe this concept?}''. The average accuracies for concept-description alignment in concrete and abstract concepts are 96.2\% and 95.2\%, respectively, resulting in an overall accuracy of 95.9\%.
For concept-image alignment, the accuracies for concrete and abstract concepts are 97.7\% and 97.2\%, respectively, with an overall accuracy of 97.5\%, indicating the high quality of our M\textsuperscript{2}ConceptBase.
%
%
To validate our cross-modal grounding double-check mechanism, we assessed grounding accuracy at each stage. Table~\ref{table:grounding} shows initial image pairing error rates of 3.0\% for concrete concepts and 15.7\% for abstract concepts in the first check. In the second check, error rates improved to 2.7\% and 8\% respectively, enhancing overall cross-modal alignment accuracy from 96.5\% to 97.3\%. These results affirm the effectiveness of our double-check mechanism.

\section{Experiments}
In this section, we present extensive experiments to demonstrate the practical application of M\textsuperscript{2}ConceptBase. We utilize M\textsuperscript{2}ConceptBase as a knowledge resource to enhance the OK-VQA~\cite{marino2019ok} (Outside Knowledge Visual Question Answering) task, leveraging its concept descriptions to boost VQA model performance (\S~\ref{subsec:4.1}--\S~\ref{subsec:4.4}). Additionally, we conduct experiments to showcase how M\textsuperscript{2}ConceptBase significantly enhances the fine-grained concept understanding abilities of retrieval-augmented MLLMs (\S~\ref{subsec:4.5}).


\subsection{Experimental Settings}
\label{subsec:4.1}

\vspace{0.5ex}
\noindent \textbf{Dataset.}
The OK-VQA dataset~\cite{marino2019ok} stands out as a comprehensive knowledge-based VQA benchmark. It comprises a collection of 14,031 diverse images paired with 14,055 thoughtfully curated questions. Distinctively, each question is crafted to necessitate external knowledge for accurate responses. The training and test set encompasses 9K and 5K image-question pairs, respectively.

\vspace{0.5ex}
\noindent \textbf{Evaluation.} Following ~\cite{tiong2022plug}, we obtain the answer by open-ended generation and perform evaluation based on exact matching. We follow previous work~\cite{tiong2022plug} and report the soft-accuracy~\cite{goyal2017making} results for the OK-VQA task.


\begin{figure}[t]
\centerline{\includegraphics[width=0.9\columnwidth]{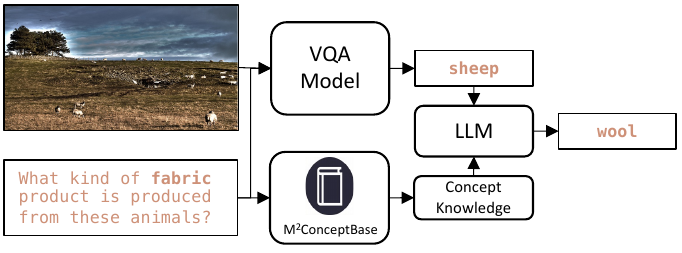}}
\caption{Illustration of OK-VQA method equipped with M\textsuperscript{2}ConceptBase and LLM.}
\label{fig:kgllm}
\end{figure}

\subsection{Methods}
\label{subsec:4.2}

\vspace{0.5ex}
\noindent \textbf{Baselines.}
We compare our approach with following baselines:
(1) \textbf{MetaLM}~\cite{hao2022language} is a semi-causal language model which supports interactions with various foundation models.
(2) \textbf{FewVLM}~\cite{jin2021good} is a low-resource prompt-based learning method for vision-language models.
(3) \textbf{PICa}~\cite{yang2022empirical} is a few-shot VQA method prompting GPT-3 with textual descriptions.
(4) \textbf{Flamingo}~\cite{alayrac2022flamingo} is a visual language foundation model with in-context few-shot learning capabilities.
(5) \textbf{PNP\_VQA}~\cite{tiong2022plug} is a zero-shot training-free modular framework composed of an image-question matching model and a captioning model.
(6) \textbf{BLIP2}~\cite{li2023blip} is a visual language foundation model that bootstraps language-image pre-training with frozen image encoders and LLMs.

\vspace{0.5ex}
\noindent \textbf{Our approach.}
We devise a simple plug-and-play module that employs our knowledge base to enhance the performance of existing OK-VQA models with an LLM as refiner. To curate an available knowledge source, we first employ an off-the-shelf image tagging tool to identify object tags in all test images and retrieve relevant concept descriptions from M\textsuperscript{2}ConceptBase. Then, as illustrated in Figure~\ref{fig:kgllm}, for each test sample, we first leverage a vanilla VQA model to get an original answer. Second, we identify the concept in the question and the answer from the vanilla VQA model. By integrating descriptions of these concepts with the output of a vanilla VQA model, we propose a knowledge-guided prompting approach to instruct an LLM to refine answers. Specifically, the question, original answer and the retrieved descriptions of concepts are provided as reference knowledge in a meticulously designed instruction to query an LLM to refine the answers. The detailed instruction is shown in Table~\ref{table:KG_LLM_VQA}.

\begin{table}[ht]
\raggedright
\begin{tabular}{p{0.9\linewidth}}
\toprule[1pt]
\textbf{OK-VQA Instruction} \\
\midrule[1pt]
Your task is to reanswer the following question based on the original answer:
\begin{itemize}
    \item \textbf{Question:} \emph{\{question\}}
    \item \textbf{Original Answer:} \emph{\{answer\}}
\end{itemize}
Here is some concept knowledge you can refer to:
\begin{itemize}
    \item The answer contains the following concepts:
         \emph{\{concept\_descriptions\_answer\}}
\end{itemize}

\begin{itemize}
    \item The question contains the following concepts:
         \emph{\{concept\_descriptions\_question\}}
\end{itemize}
\textbf{Hint:} If you think the original answer is incorrect based on the concept knowledge, try to give the correct answer directly. If it is correct, just repeat the original answer. \\
\textbf{Output Format:} A short answer, no explanation, no other output. \\
\textbf{Your Answer:} \\
\bottomrule[1pt]
\end{tabular}

\caption{Instruction of OK-VQA method equipped with M\textsuperscript{2}ConceptBase and LLM.}
\label{table:KG_LLM_VQA}
\end{table}

\subsection{Implementation Details}
\label{subsec:4.3}

Following Yang et al.~\cite{yang2022empirical}, we use vinVL~\cite{zhang2021vinvl} as image tagging tools. For VQA models, we choose PNP\_VQA~\cite{tiong2022plug} and BLIP2~\cite{li2023blip} as our backbones. Specifically, we use the base version of PNP\_VQA, which is BLIP~\cite{li2022blip} with the ViT-L/16  for the image-question matching module and UnifiedQAv2~\cite{khashabi2022unifiedqa} for the question-answering module. For BLIP2, the vision encoder is ViT-g from EVA-CLIP~\cite{sun2023eva}, and the LLM is FlanT5\textsubscript{XL}~\cite{chung2024scaling}. We utilize GPT-3.5-Turbo as our LLM refiner, setting max\_tokens to 5 and the temperature to 0.

\subsection{Results and Analyses}
\label{subsec:4.4}

As illustrated in Table~\ref{table:okvqa}, with the help of M\textsuperscript{2}ConceptBase, our approach positively influences the performance of two backbone models (PNP\_VQA and BLIP2), underscoring the its efficacy. Specifically, our approach achieves a performance increase of 1.6 percents over PNP\_VQA and 0.4 percents over BLIP2, surpassing previous state-of-the-art models. As depicted in Figure~\ref{fig:kgllm}, our approach leverages knowledge description retrieved with the key concept ``fabric'' from the question and the original VQA output ``sheep''.  By referencing the description, the LLM reanalyzes the question's intent and corrects the original output, providing the accurate answer ``wool''. This enhancement can be attributed to the descriptive knowledge associated with ``sheep'' and ``fabric''.


\begin{table}[t]
\centering
\resizebox{0.7\linewidth}{!}{%
\begin{tabular}{l p{0.3\columnwidth}}
\toprule
\textbf{Method} & \textbf{Accuracy (\%)} \\
\midrule
MetaLM & 11.4 \\
FewVLM\textsubscript{base} & 11.6 \\
FewVLM\textsubscript{large} & 16.5 \\
PICa\textsubscript{base} & 16.4 \\
PICa\textsubscript{full} & 17.7 \\
Flamingo\textsubscript{3B} & 41.2 \\
\hline
PNP\_VQA\textsubscript{base} & \underline{23.2} \\
Ours\textsubscript{PNP\_VQA\textsubscript{base}} & \textbf{24.8(+1.6)} \\
\hline
BLIP2\textsubscript{flant5xl} & \underline{41.1} \\
Ours\textsubscript{BLIP2\textsubscript{flant5xl}} & \textbf{41.5(+0.4)} \\
\bottomrule
\end{tabular}
}
\caption{Zero-shot OK-VQA results.}
\label{table:okvqa}
\end{table}

\begin{table}[t]
\centering
\resizebox{0.7\linewidth}{!}{%
\begin{tabular}{l p{0.25\columnwidth}}
\toprule[1pt]
\textbf{Method} & \textbf{Accuracy(\%)}\\
\midrule[1pt]
PICa & 9.9 \\
PICa+Richpedia & 9.9 \\
PICa+ConceptNet & 5.8 \\ 
\textbf{PICa+M\textsuperscript{2}ConceptBase} &\textbf{11.7} \\
\bottomrule[1pt]
\end{tabular}%
}
\caption{Ablation study with other KBs for OK-VQA.}
\label{table:other_kb}
\end{table}

As shown in Table~\ref{table:other_kb}, we compare our knowledge base (KB) with other KBs using PICa\textsubscript{base} as the backbone and test 200 samples from the OK-VQA test set under the same settings. M\textsuperscript{2}ConceptBase includes detailed concept descriptions, whereas other multimodal knowledge bases (MMKBs) often lack such information. For example, Richpedia does not include relevant concept information and thus shows no improvement for the OK-VQA test set. We compare M\textsuperscript{2}ConceptBase with a text-based concept graph (namely ConceptNet) containing concept-related knowledge, where we concatenate all ``isA'' relations into a single sentence. The results in Table~\ref{table:other_kb} demonstrate that our MMKB contains richer conceptual descriptive knowledge compared to ConceptNet, resulting in improved performance when using PICa as the backbone.

\subsection{Retrieval Augmented MLLMs}
\label{subsec:4.5}

Given the notable deficiencies of MLLMs in grasping fine-grained, long-tail concepts, and the wealth of cross-modal aligned knowledge provided by M\textsuperscript{2}ConceptBase, we devised two retrieval augmented generation (RAG) tasks to showcase the potential of retrieval augmented MLLMs' cross-modal concept understanding capabilities. Specifically, we assess the performance of MLLMs in two subtasks: zero-shot visual concept recognition (0-shot VCR) and visual concept knowledge description generation (VCKDG) in in-context learning settings. For 0-shot VCR task, MLLMs are queried by the question of ``What is it?'' to recognize the specific concept depicted in the input image. While in the VCKDG task, MLLMs are provided with the image along with the prompt ``Based on your understanding of the visual concept in the image, provide a detailed introduction to the background knowledge related to the concept.'', to generate a detailed description of the concept.



\begin{table}[htbp]
\centering
\small
\resizebox{0.9\columnwidth}{!}{
\begin{tabular}{p{0.85\columnwidth}}
\toprule
\textbf{The List of 30 Fine-Grained Visual Concepts} \\
\midrule
blueberry jam, LCD TV, pedilanthus tithymaloides,  \\
purple root orchid, Orlistat tablets, shipborne helicopters,  \\
budgerigar, sedum lineare, spicy lotus root slices, gelsemium elegan, \\
professional racers, license plate number, self-tapping snails, \\
ammonium bicarbonate, hexagonal bottle, pugs, card reader, \\
fried silver fish with eggs, electric kettle, hornet, chocolate cake,\\
cold welding machine, sulfur, short boots, Guqin, \\
pull rod bag, Deer Antler, sugar painting, sponge, lotus \\
\bottomrule
\end{tabular}
}
\caption{Fine-grained visual concepts in our dataset.}
\label{table:concept_list}
\end{table}

\vspace{0.5ex}
\noindent \textbf{Dataset.} We curate a concept image dataset, which encompasses a total of 200 images searched from Internet. It covers 30 fine-grained visual concepts from M\textsuperscript{2}ConceptBase (see the detailed concept list in  Table~\ref{table:concept_list}), where each concept has an average of 6.67 test images. 

\vspace{0.5ex}
\noindent \textbf{Evaluation.} 
For 0-shot VCR task, we calculate accuracy based on the exact match of the concept name. For VCKDG task, we employ a strategy using an LLM (\emph{i.e.} GPT-3.5-Turbo) as a judge to compare the win rate of MLLMs' responses between the RAG and non-RAG settings.

\vspace{0.5ex}
\noindent \textbf{Experimental Setup.} For each task, we test three MLLMs in both the zero-shot (non-RAG) and retrieval-augmented generation (RAG) settings. Specifically, for the non-RAG setting, we just input the image and the query prompt to get a response from MLLMs. For the RAG setting, we extract up to ten images and one concept description for every concept from M\textsuperscript{2}ConceptBase to construct a multimodal knowledge base. We utilize the OpenCLIPEmbeddingFunction API in chromadb~\footnote{https://www.trychroma.com/} to construct a vector database for all images. In the case of MLLMs with the RAG setting, we initially retrieve the most similar image from the vector database and consider its concept label as the predicted concept for the input image. Subsequently, we integrate the retrieved concept or description into the query prompt to obtain a response with the RAG prompt of ``The retrieved result shows a \{\emph{concept\_name}\} in the image. What is it?'' and ``According to the retrieved conceptual knowledge: \{\emph{concept\_name}\}: \{\emph{concept\_description}\}. Please provide a detailed introduction to the background knowledge related to the visual concept.'', respectively.

\vspace{0.5ex}
\noindent \textbf{Baselines.}
We compare the performance of three MLLMs in both RAG and non-RAG settings: (1) \textbf{VisualGLM}~\cite{du2022glm, ding2021cogview}: VisualGLM is an open-source multimodal dialogue language model based on ChatGLM that supports images, Chinese, and English. (2) \textbf{QWen-VL}~\cite{bai2023qwen}: QWen-VL, Alibaba Cloud’s visual multimodal version of the QWen large model series, accepts inputs of images, text, and bounding boxes, and outputs text and bounding boxes. (3) \textbf{GPT-4V}~\cite{achiam2023gpt}: GPT-4V, the most powerful LLM from OpenAI, can process both images and text, generating textual outputs.

\begin{table}[t]
\centering
\resizebox{\columnwidth}{!}{
\begin{tabular}{lcccc}
\toprule
&  & \multicolumn{3}{c}{\textbf{VCKDG}} \\
\cmidrule{3-5}
\textbf{Method} & \textbf{0-shot VCR Acc(\%)} & \multicolumn{1}{c}{\textbf{Win$\uparrow$}} & \multicolumn{1}{c}{\textbf{Lose$\downarrow$}} & \multicolumn{1}{c}{\textbf{Tie$\leftrightarrow$}}\\
\midrule
VisualGLM(w. RAG) & \textbf{35.0} &  &  & \\
vs. VisualGLM(w.o. RAG) & 12.0 & \textbf{82.0\%} & 8.0\% & 10.0\%\\ 
\midrule
QWen-VL(w. RAG) & \textbf{89.0} &  &  & \\
vs. QWen-VL(w.o. RAG) & 29.0 & \textbf{76.5\%} & 16.0\% & 7.5\%\\ 
\midrule
GPT-4V(w. RAG) & \textbf{83.0} &  &  & \\
GPT-4V(w.o. RAG) & 20.5 & \textbf{56.0\%} & 36.5\% & 7.5\%\\ 
\bottomrule
\end{tabular}
}
\caption{Results for 0-shot VCR and VCKDG tasks.}
\label{table:mmrag}
\end{table}

\begin{figure}[t]
\centerline{\includegraphics[width=0.85\columnwidth]{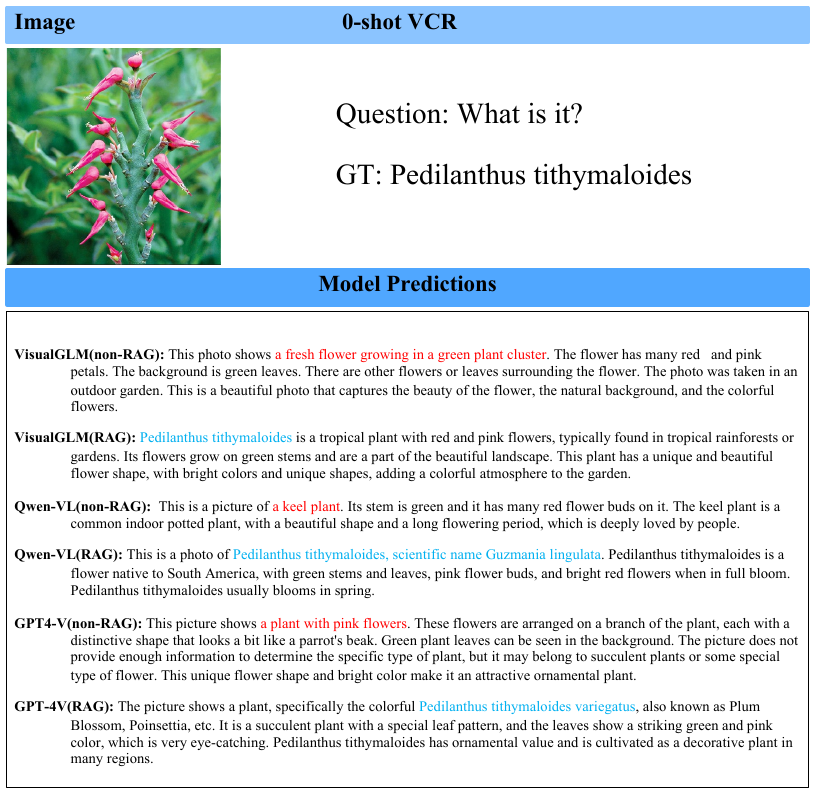}}
\caption{Case study for 0-shot VCR task.}
\label{fig:case_study}
\end{figure}

\vspace{0.5ex}
\noindent \textbf{Results and Analyses.} 
In Table~\ref{table:mmrag}, we observe consistent performance enhancements across all MLLMs in both the 0-shot VCR and VCKDG tasks when retrieval augmentation with the multimodal knowledge base from M\textsuperscript{2}ConceptBase is applied. This underscores the significant contribution of M\textsuperscript{2}ConceptBase in augmenting MLLMs' visual concept understanding in RAG scenarios. Specifically, VisualGLM, QWen-VL, and GPT-4V achieve improvements of 23\%, 60\%, and 62.5\% , respectively, in the 0-shot VCR task. In the VCKDG task, MLLMs with RAG setting demonstrated a notable win rate against non-RAG setting in pairwise comparisons with LLM (\emph{i.e.}, GPT-3.5-turbo). As illustrated in Figure~\ref{fig:case_study}, we present a qualitative example for the 0-shot VCR task showcasing the responses of each MLLM. Notably, for the fine-grained concept ``Pedilanthus tithymaloides'', all MLLMs, including GPT-4V, provide incorrect answers or produce coarse-grained or hallucinative descriptions for the visual concept (\emph{e.g.}, ``a fresh flower'' for VisualGLM, ``a keel plant'' for QWen-VL, ``a plant with pink flowers'' for GPT-4V). In contrast, retrieval-augmented MLLMs generated correct answers with more accurate descriptions.

\section{Related Work}

Recently, Multimodal Large Language Models (MLLMs) have shown great promise in various multimodal tasks. Previous research~\cite{petroni2019language, alkhamissi2022review} indicates that large language models store extensive knowledge within their parameters through the pre-training stage. Additionally, MLLMs implicitly encode abundant cross-modal knowledge by pretraining on image-text corpora. However, long-tail knowledge is typically neglected due to its low frequency in the pre-training corpus. In contrast, MMKBs provide explicit long-tail aligned knowledge, which can serve as valuable supplements to MLLMs.

Current MMKBs are typically constructed through two main approaches: collecting images for textual knowledge bases~\cite{ferrada2017imgpedia, liu2017robust, wang2020richpedia, alberts2020visualsem, zhang2023aspectmmkg}, or extracting visual knowledge directly from images~\cite{chen2013neil, li2020gaia, wen2021resin, krishna2017visual}. The former method involves searching for candidate images based on textual entities or concepts and then employing a filtering mechanism to obtain matched images. The latter method utilizes automated techniques such as object detection, image tagging, and visual relationship extraction to extract entity and concept information, although manual efforts may also be employed for annotating knowledge bases. Notably, these existing MMKBs are primarily entity-centric, in contrast to M\textsuperscript{2}ConceptBase, which stands out as the first concept-centric MMKB.


\section{Conclusion}

Given the importance and scarcity of concept-centric multimodal knowledge bases (MMKBs), we propose a novel context-aware multimodal symbol grounding framework to construct them from large-scale image-text pairs and encyclopedias. Utilizing this framework, we have developed M\textsuperscript{2}ConceptBase, the first concept-centric MMKB, comprising 951,089 images and 151,776 concepts. Each concept in M\textsuperscript{2}ConceptBase is associated with an average of 6.27 relevant images and a detailed description. Data quality experiments demonstrate over 95.5\% accuracy in alignment between concepts, descriptions, and images, confirming its high quality. 
Experimental results on the downstream task, namely OK-VQA, validate the effectiveness of M\textsuperscript{2}ConceptBase in enhancing visual comprehension, underscoring its value and usefulness. Moreover, the significant enhancement of fine-grained concept understanding in MLLMs through 0-shot VCR and VCKDG tasks further demonstrates the substantial contribution of M\textsuperscript{2}ConceptBase.


\begin{acks}
This work is supported by the National Natural Science Foundation of China (No.62072323, U21A20488, No.62102276), Shanghai Science and Technology Innovation Action Plan (No.22511104700), Zhejiang Lab Open Research Project (No.K2022NB0AB04).
\end{acks}



%

\bibliographystyle{ACM-Reference-Format}
\bibliography{sample-base}

\appendix

\end{document}